\documentclass[a4paper,10pt,twocolumn]{article}
\usepackage[utf8]{inputenc}
\usepackage{multicol,graphicx}
\usepackage{xcolor}
\usepackage{subcaption}
\usepackage{booktabs}
\usepackage{mathptmx}
\usepackage[T1]{fontenc}
\usepackage{textcomp}
\usepackage{url}
\usepackage{csquotes}
\usepackage{siunitx}
\usepackage[hidelinks]{hyperref}
\usepackage{nameref}
\usepackage{multirow}
\usepackage[T1]{fontenc}
\usepackage[english]{babel}

\usepackage[backend=bibtex,style=ieee,doi=false,isbn=false,url=true,maxnames=6,citestyle=numeric-comp,giveninits=true]{biblatex}

\fontfamily{ptm}\selectfont

\bibliography{references.bib}

\usepackage[font=small]{caption}
\usepackage{amsmath}

\usepackage{cleveref}

\makeatletter
\newcommand\setcurrentname[1]{\def\@currentlabelname{#1}}
\makeatother

\newcommand{\mysection}[1]{\vspace{0.4cm} \uppercase{#1}\setcurrentname{#1}\phantomsection \vspace{0.4cm}}
\newcommand{\mysubsection}[1]{\hspace{10pt}\textit{#1}\setcurrentname{#1}\phantomsection}

\begin{document}
	
\setlength{\textfloatsep}{10pt plus 1.0pt minus 2.0pt}	
\setlength{\columnsep}{1cm}





\twocolumn[%
\begin{@twocolumnfalse}
\begin{center}
	{\fontsize{14}{18}\selectfont
       \textbf{\uppercase{
       Synthesizing EEG Signals from Event-Related Potential Paradigms with Conditional Diffusion Models}}\\}
    \begin{large}
        \vspace{0.6cm}
        Guido Klein\textsuperscript{1$\star$},
        Pierre Guetschel\textsuperscript{1$\star$}, 
        Gianluigi Silvestri\textsuperscript{1,2},
        Michael Tangermann\textsuperscript{1}\\
        \vspace{0.6cm}
        \textsuperscript{$\star$}These authors contributed equally to this work.\\
        \textsuperscript{1}Donders Institute for Brain, Cognition and Behaviour,
        Radboud University, Nijmegen, Netherlands\\
        \textsuperscript{2}OnePlanet Research Center, imec-the Netherlands, Wageningen, Netherlands\\
       \vspace{0.5cm}
        E-mail: \href{mailto:pierre.guetschel@donders.ru.nl}{pierre.guetschel@donders.ru.nl}
        \vspace{0.4cm}
    \end{large}
\end{center}	
\end{@twocolumnfalse}%
]%


ABSTRACT: 

Data scarcity in the brain-computer interface field can be alleviated through the use of generative models, specifically diffusion models. While diffusion models have previously been successfully applied to electroencephalogram (EEG) data, existing models lack flexibility w.r.t.~sampling or require alternative representations of the EEG data. To overcome these limitations, we introduce a novel approach to conditional diffusion models that utilizes classifier-free guidance to directly generate subject-, session-, and class-specific EEG data. In addition to commonly used metrics, domain-specific metrics are employed to evaluate the specificity of the generated samples. The results indicate that the proposed model can generate EEG data that resembles real data for each subject, session, and class. 

\mysection{Introduction}
\label{section:introduction}
\\
One of the most significant challenges of data scarcity in the brain-computer interface (BCI) field is that the acquisition of annotated data is a time-intensive endeavor. The lack of large labeled datasets can be a bottleneck for many machine learning algorithms~\cite{EvolvingDeepLearning}. Additionally, class imbalances typically found in event-related potential (ERP) protocols which are among the most commonly used EEG-BCI paradigms~\cite{leeEEGDatasetOpenBMI2019}, can be detrimental to classifier performance. Moreover, multiple populations are underrepresented in the current corpus of EEG data~\cite {choySystemicRacismEEG2021, EvolvingDeepLearning}.

Generative models offer a promising solution to alleviate this data scarcity. Diffusion models, in particular, have shown the ability to generate high-quality data in a variety of domains, including images~\cite{dhariwalDiffusionModelsBeat} and audio~\cite{huangMakeAnAudioTextToAudioGeneration2023}. Current implementations of diffusion models for EEG data generation are either trained directly on EEG data or use an alternative representation, such as electrode frequency density maps~\cite{tosatoEEGSyntheticData2023}, spatial covariance matrices~\cite{juScoreBasedDataGeneration2023}, time-frequency maps, and latent representations~\cite{aristimunhaSyntheticSleepEEG, sharmaMEDiCMitigatingEEG, zhouGenerativeAIEnables2023}. 

Models trained on alternative representations, while potentially being easier to train, require an additional pre- and post-processing step, which can hinder their usability. The models trained directly on EEG data are either unconditioned~\cite{tormaEEGWaveDenoisingDiffusion2023,vetterGeneratingRealisticNeurophysiological2023}, which means that the samples are always generated from the full data distribution, or conditioned, which means that the models are trained on the full data distribution but samples can be generated from a selected part of the data distribution. This conditioning can either be achieved using a classifier~\cite{dhariwalDiffusionModelsBeat, wangDiffMDDDiffusionBasedDeep2024} or by \textit{classifier-free guidance}~\cite{hoClassifierFree, shuDataAugmentationSeizure2023}, which achieves conditioning without the need for a noisy classifier.

Despite the capability of diffusion models to generate high-quality EEG data, there is a lack of proper metrics to quantify the quality of the generated samples. Currently used metrics are either adopted from the image domain, are domain-invariant, or rely on classifier performance~\cite{habashiGenerativeAdversarialNetworks2023, neifarDeepGenerativeModels2023, tormaEEGWaveDenoisingDiffusion2023, aristimunhaSyntheticSleepEEG, juScoreBasedDataGeneration2023, wangDiffMDDDiffusionBasedDeep2024, zhouGenerativeAIEnables2023, sharmaMEDiCMitigatingEEG, shuDataAugmentationSeizure2023, tosatoEEGSyntheticData2023}. The metrics from the image domain, the Fr\'echet inception distance (FID)~\cite{dhariwalDiffusionModelsBeat} and the inception score (IS)~\cite{dhariwalDiffusionModelsBeat}, rely on the activations and output of a standardized trained neural network called Inception V3~\cite{dhariwalDiffusionModelsBeat}. Unfortunately, there is no universally adopted trained network for EEG data, which makes fair and reliable comparison impossible. Additionally, the domain-invariant metrics are incapable of discerning which domain-relevant features are generated well by the model. For example, if there is a low Euclidean distance between the generated and real samples, then that is likely due to a similarity in multiple domain-relevant features, such as amplitude and peak latencies, this makes it unclear which domain-specific features are properly generated. Similarly, metrics based on classifier performance are also unable to disentangle these features. Hence, there is a need for a set of metrics that can capture these domain-specific features.

\mysubsection{Research questions and objectives.}

The following research question is investigated: Can we generate artificial ERP examples that are specific to a subject, session, and class using conditional diffusion models with classifier-free guidance?

\begin{figure*}[h!]
\centering

\begin{subfigure}[t]{.52\linewidth}
    \centering
	\includegraphics[width=\linewidth]{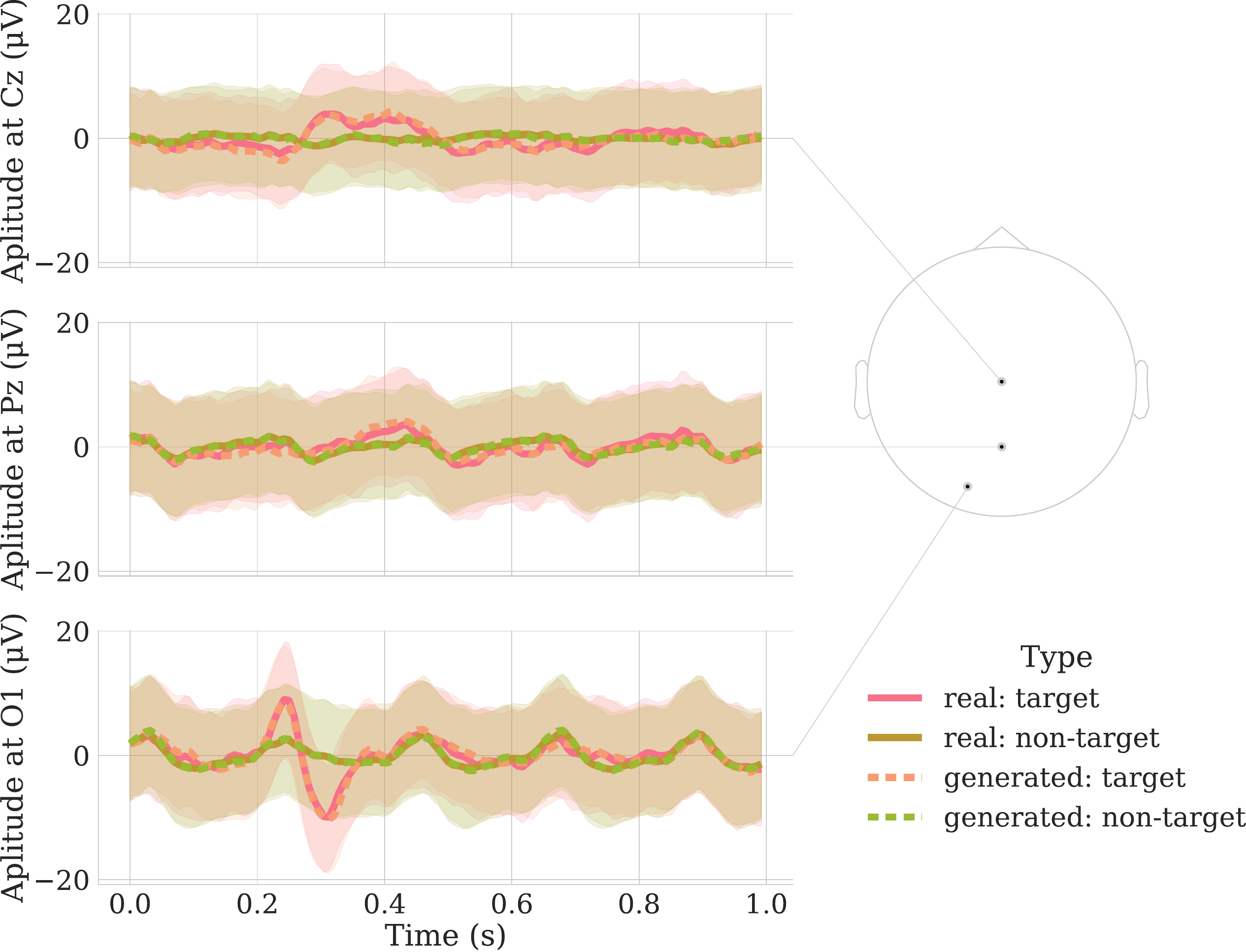}
	\caption{Real and generated average temporal \textbf{ERP} responses of target and non-target for three selected EEG channels. The error bands indicate the standard deviations of the data.}
	\label{fig:timecourse}
\end{subfigure}\hfill
\begin{subfigure}[t]{.42\linewidth}
	\centering
	\includegraphics[width=\linewidth]{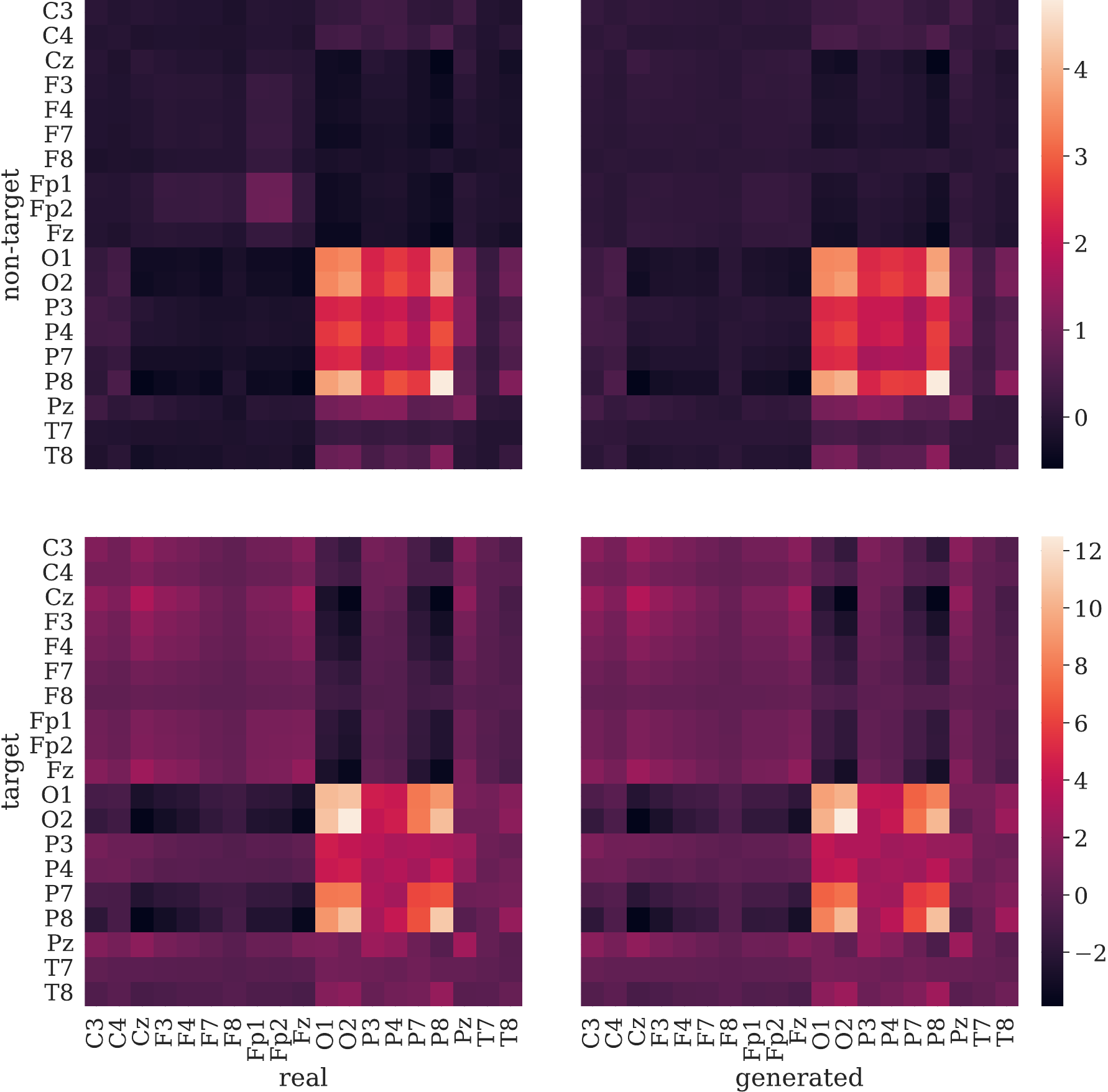}
	\caption{
        \textbf{Covariance matrices} of averaged
         real (left) and generated (right) responses for both non-target (top) and target (bottom) responses.}
	\label{fig:SCM}
\end{subfigure}
\caption{Figures (\subref{fig:timecourse}) and (\subref{fig:SCM}) provide two different comparisons between real and generated data. The figures are based on the averages of EEG data (522 target and 2875 non-target examples), which are sampled for the combination of subject and session combination that resulted in the worst ABA metric (subject~52, session~1). The model with the highest ABA metric over all subjects and sessions (i.e. 600k training steps) was used to generate the samples.}
\label{fig:data_viz}
\end{figure*}

To answer this question, we train a novel conditional diffusion model to flexibly generate each combination of conditions, i.e., subject, session, and class. An example of generated data can be found in \autoref{fig:data_viz}. Domain-invariant and image-domain metrics are used to evaluate the quality of generated samples during training. However, as previously noted, these metrics are unable to capture domain-specific features, which makes it impossible to evaluate the ability of the model to generate ERP data that is specific to a certain condition. Therefore, we aim to introduce domain-specific metrics, which use ERP-related features, to verify the ability of the model to generate ERP data that is specific to each combination of conditions. 

\mysection{Data Description} 
\\
The conditional diffusion model is trained on a visual ERP dataset collected by Lee and colleagues~\cite{leeEEGDatasetOpenBMI2019}. Visual ERP responses are elicited using a modified oddball paradigm. The most prominent ERP feature is expected to be a positive deflection that occurs approximately 300\,ms (referred to as a P300) after being presented with a relatively infrequent target stimulus following multiple non-target stimuli~\cite{townsendNovelP300basedBraincomputer2010}. This dataset is one of three datasets that were recorded to study BCI-inefficiency across three major BCI paradigms: visual ERP, motor imagery, and steady-state visually evoked potential protocols~\cite{leeEEGDatasetOpenBMI2019}.  In the study, fifty-four participants underwent two sessions which were held on different days~\cite{leeEEGDatasetOpenBMI2019}. 

During data recording, each visual ERP session was divided into a train and test run. During the train run, trials were not decoded, while during the test run, the trials were decoded and feedback was given to the participant after each trial~\cite{leeEEGDatasetOpenBMI2019}. Both runs of the same session are combined to train the diffusion model, as they employ the same copy-spelling tasks.

The dataset has been obtained and preprocessed using the Mother of All BCI Benchmarks~\cite{aristimunhaMotherAllBCI2023}. It was preprocessed with a relatively simple pipeline. First, 19 channels (Fp1, Fp2, F7, F8, F3, F4, Fz, T7, T8, C3, C4, Cz, P7, P8, P3, P4, Pz, O1, and O2) were selected that provide full scalp coverage. Secondly, the data was bandpass filtered between 1 and 40\,Hz with a 4th-order Butterworth filter. Thirdly, the data was downsampled from 1000\,Hz to 128\,Hz. Fourthly, epochs were constructed as 1-second windows, starting from the stimuli onsets. Lastly, peak-to-peak epoch rejection is applied with a threshold of 150\,\si{\micro\volt}. This removed 14.5\,\% of epochs, equivalent to 63672~epochs out of a total of 438840~epochs. Subject~17 was dropped due to excessive artifacts.

\mysection{Diffusion Models}
\\
Diffusion models are a generative modeling paradigm where data is progressively destroyed by injecting Gaussian noise, and a neural network is trained to reverse this process~\cite{hoClassifierFree}. Song and colleagues provide a continuous formulation of such a process, formulated as a stochastic differential equation (SDE), and show how a neural network can be implemented to learn the reverse SDE~\cite{songScoreBasedGenerativeModeling2021}. In this work, the implementation is based on the variance persevering SDE (VP\,SDE), which is the continuous equivalent to the noise injection used in the denoising diffusion probabilistic model (DDPM)~\cite{hoDenoisingDiffusionProbabilistic, songScoreBasedGenerativeModeling2021}. Furthermore, \enquote{classifier-free guidance} is used to condition the model on the subject, session, and class in parallel~\cite{hoClassifierFree}. 

\mysubsection{Implementation details:}
The neural network to reverse the destruction of the data is based on the architectures introduced by Torma et~al. and Shu et~al., called EEGWave and diff-EEG~\cite{tormaEEGWaveDenoisingDiffusion2023, shuDataAugmentationSeizure2023}. Two key differences were introduced: 1) the timestep embedding was rewritten to be compatible with the VP SDE, because both models use DDPM noise injection, and 2) no normalization is applied to the EEG data. 

The model is trained for 900\,k~steps, with the model being evaluated every 100\,k~steps. An exponential moving average of the weights is used for sampling and metric calculation. This sampling is done by a predictor-corrector sampler~\cite{songScoreBasedGenerativeModeling2021}. Preliminary results indicated that an increase in the signal-to-noise ratio (SNR) of the corrector increases the amplitude of the generated EEG data. The number of generated samples matches the number of real samples available for a given subject/session/class combination. This includes the samples used to compute the validation loss but excludes samples removed by the epoch rejection.

For more information about the implementation please refer to: \url{https://neurotechlab.socsci.ru.nl/resources/generative_models/}

\mysection{Similarity Metrics}
\label{section:similarity metrics}

Metrics quantifying the similarity between generated and real data are crucial for model comparison and evaluation. These similarity metrics are divided into four categories: \textit{classifier performance}, \textit{domain-invariant}, \textit{image-domain}, and \textit{domain-specific}. This section will also discuss metric-specific baselines for interpreting the scores obtained on the domain-invariant and domain-specific metrics.

Domain-invariant and domain-specific metrics are computed between the real and generated data within one condition, i.e., a combination of subject, session, and class. These metrics have been computed separately per condition, but their average across conditions is reported.

\mysubsection{Classifier performance:} We compare the performance of a classifier trained on generated data with the performance of a classifier trained on generated data. Both conditions use the same test sets which contain only real data. 
This metric is denoted as the averaged balanced accuracy (\textbf{ABA}) and the score obtained when training on the real data is reported as the \textit{within-session baseline}. 
Specifically, training is subject-specific and implements a within-session five-fold stratified cross-validation. The classifier is a regularized least-squares Linear Discriminant Analysis (LDA)~\cite{BLANKERTZ2011814}. The LDA is trained on features that represent the average amplitude across channels within non-overlapping time windows, which span between 0.1 to 0.9 seconds and are each 0.1 seconds long.

\mysubsection{Domain-invariant metrics}\label{section:domain-invariant}, such as the sliced-Wasserstein distance (SWD), mean squared error, and Jenson-Shannon divergence, can be used to measure the (dis)similarity between the generated and real data. However, there is no consensus on which metric to use, with multiple articles implementing different domain-invariant metrics~\cite{habashiGenerativeAdversarialNetworks2023, tormaEEGWaveDenoisingDiffusion2023, sharmaMEDiCMitigatingEEG, aristimunhaSyntheticSleepEEG}. Therefore, we arbitrarily choose to implement the SWD~\cite{rabinWassersteinBarycenterIts2012}. The domain-specific metrics introduced later allow for a more nuanced approach to metric selection.

\mysubsection{Image-domain metrics} require a pre-trained classifier, as they are based on its latent activations. Inception~V3~\cite{dhariwalDiffusionModelsBeat} is the one used for images but is not suitable for EEG data, so we have trained an EEGNet architecture~\cite{lawhernEEGNetCompactConvolutional2018} for this purpose. This trained EEGNet model has been made public, allowing future studies to calculate the FID and IS using the same model\footnote{EEGNet checkpoint: \url{https://huggingface.co/guido151/EEGNetv4}}.

Although the IS is a commonly used metric from the image domain, it is not included in our analysis. The IS leverages the outputs of the classifier to extract information about the quality and diversity of the samples. However, the diversity of samples generated by a conditional diffusion model is, practically speaking, arbitrary. Moreover, the trained EEGNet is biased toward the majority class. Thus, the output of the classifier is unable to accurately measure the diversity of the samples. Therefore, we have decided to exclude it from the analysis.

Fortunately, the other important metric from the image domain, the FID, does not suffer from the same problem as the IS. The FID is calculated by computing the Fr\'echet distance between a Gaussian fitted to the mean and standard deviations of the activations in the last pooling layer of a trained classifier in response to the real and generated data~\cite{heuselGANsTrainedTwo2018}. The Fr\'echet distance is small when the features that are picked up by the trained classifier are similar. Unfortunately, it is almost impossible to discern which features contribute to a low FID, because it is unclear which features would cause similar activations. This problem, however, can be tackled by the domain-specific metrics.

\mysubsection{Four Domain-specific metrics} are introduced in the following section. They are designed to address the shortcomings of the previous metrics to evaluate the ability of the model to generate domain-specific features. In particular, their design exploits the stationarity of ERP brain responses by using averaged responses, thus increasing the SNR~\cite{BLANKERTZ2011814}. 

The difference between generated and real P300 peaks is measured by the \textit{peak latency delta} (\textbf{PLD}) and \textit{peak amplitude delta} (\textbf{PAD}) metrics. These metrics only consider the channel with the most prominent P300 peak when averaged over all the real target data, which is channel "O1" in the case of the Lee 2019 ERP dataset. The PAD is computed by taking the absolute difference in \si{\micro\volt} between the highest peak in the real and generated data at the selected channel. The PLD is the absolute difference in time offset between these peaks, measured in \si{ms}.  Only target trials are included, as a P300 peak is necessary to compute these metrics.

One of the main downsides of using averaged data is that information about the diversity of samples is lost. Nonetheless, it is important that the model can generate a variety of EEG data. The final metric, called the \textit{standard deviation Manhattan distance} (\textbf{SD-MD}), addresses this concern by computing the absolute differences in standard deviation values for both real and generated data for each channel and subsequently averaging these differences over all channels.

\mysubsection{Between-session variability} 
is reported as a baseline for the domain-invariant and domain-specific metrics. This is achieved by measuring the variability of two sessions of the same subject and the same class. Assuming that the model performs better than the between-session variability, and assuming that variability between subjects is larger than the between-session variability within the same subject and the same class, this test checks if the model can generate data that is specific to only one combination of subject, session, and class.

\mysection{Results}

One diffusion model is trained and conditioned on all combinations of subjects, sessions, and classes simultaneously. Ideally, this training approach should allow the model to generate data that is specific to a subject, session, and class. 

\begin{figure*}[h!]
	\centering
	\includegraphics[width=\linewidth]{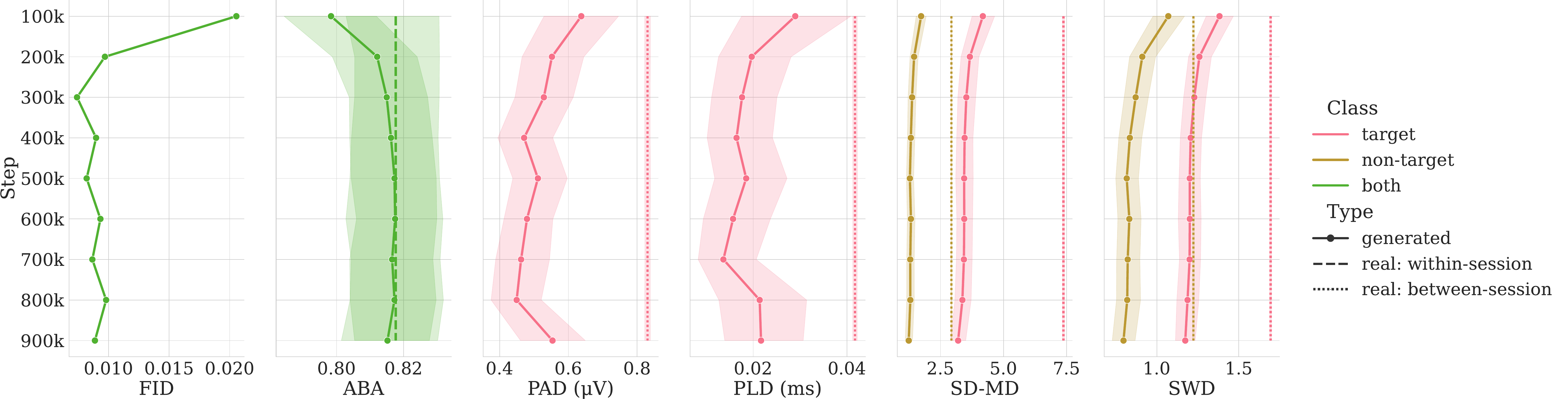}
	\caption{Scores of every metric measured on the generated data of the model at multiple training steps. The \enquote{real: within-session} baseline is computed by taking the ABA metric on the real data within a session of the same participant. The \enquote{real: between-session} baseline displays the variance between two sessions of the same participant on that particular metric. The band shows the 95\,\% confidence interval.
 }
	\label{fig:Metrics}
\end{figure*}

The model's ability to generate data for all combinations is assessed every 100k~training steps using the metrics introduced in \nameref{section:similarity metrics}. The results are visualized in~\autoref{fig:Metrics}. The model of the training step that achieves the highest ABA is further evaluated using the other metrics. These outcomes are provided in \autoref{tab:metrics}. Additionally, high-resolution plots were created for the combination of subject 52 and session 1, as this combination resulted in the worst generated data according to the ABA metric. The plots of this combination are shown in~\autoref{fig:data_viz}. 

\mysubsection{Trends in training and model performance:}
In general, the model outperforms the between-session variability in the domain-invariant (SWD), and domain-specific (PAD, PLD, and SD-MD) metrics. Using the ABA metric, it performs similarly to the within-session baseline. There seems to be a common trend in the scores obtained by the model. Namely, all metrics show a relatively large increase in performance between 100k and 200k~training steps and are relatively stable afterward (see~\autoref{fig:Metrics}). The ABA and PAD do show a slight decrease in performance between the 800k and 900k~training steps, but this decrease is not large enough to conclude that the overall performance is decreasing.

\mysubsection{Classifier performance:}
The ABA is 0.818 for the within-session real-data baseline. For generated samples, the highest ABA is 0.817, which is achieved by samples from the model at 600k~training steps. 

According to ABA, the best subject/session combination is subject~51 in session~2, for which the generated data outperforms the within-session baseline by 0.045. Conversely, the worst combination is subject~52 in session~1, for which the classification score using generated data decreases by 0.043 compared to the within-session baseline. Thus, even in the worst case, the generated data has LDA features that are very similar to the real data.

\begin{table}[h!]
\footnotesize 
    \caption{Scores of every metric measured on data sampled from the best checkpoint according to the ABA metric compared to a baseline, provided that there is a baseline. The baseline for the ABA is within-subject and within-session, while the baselines for the SWD, PAD, PLD, and SD-MD are computed using the variance between the two sessions of the same subject} 
    \label{tab:metrics}
    \begin{tabular*}{1\columnwidth}{l @{\extracolsep{\fill}} cccccc}
    \toprule
    & \multicolumn{3}{c}{Generated} & \multicolumn{3}{c}{Baseline}\\
    \cmidrule(r{2ex}){2-4}\cmidrule(lr{2ex}){5-7}
         & target & non-tgt. & both & target & non-tgt. & both \\
         \midrule
         FID $\downarrow$   & - & - & $93e^{-4}$ & - & - & - \\
         ABA $\uparrow$   & - & - & 0.817 & - & - & 0.818 \\
         PAD $\downarrow$   & 0.48 & - & - & 0.83 & - & - \\
         PLD $\downarrow$   & 0.016 & - & - & 0.042 & - & - \\
         SD-MD $\downarrow$   & 3.44 & 1.33 & 2.39 & 7.38 & 2.94 & 5.16 \\
         SWD $\downarrow$   & 1.20 & 0.83 & 1.02 & 1.69 & 1.22 & 1.46 \\
         \bottomrule
    \end{tabular*}
\end{table}

\mysubsection{Similar results on SWD and SD-MD:}
The domain-invariant SWD of the generated data is better than its between-session baseline. Interestingly, the domain-specific SD-MD shows a strikingly similar pattern. Both with target and non-target data, the offset between the generated data and their respective baseline are almost identical. Thus, it seems that the SWD captures some information about the standard deviation, as this is explicitly measured by the SD-MD. The standard deviation of three channels for the worst subject, according to ABA, can be seen in \autoref{fig:data_viz}.

\begin{figure}[h!]
	\centering
	\includegraphics[width=\linewidth]{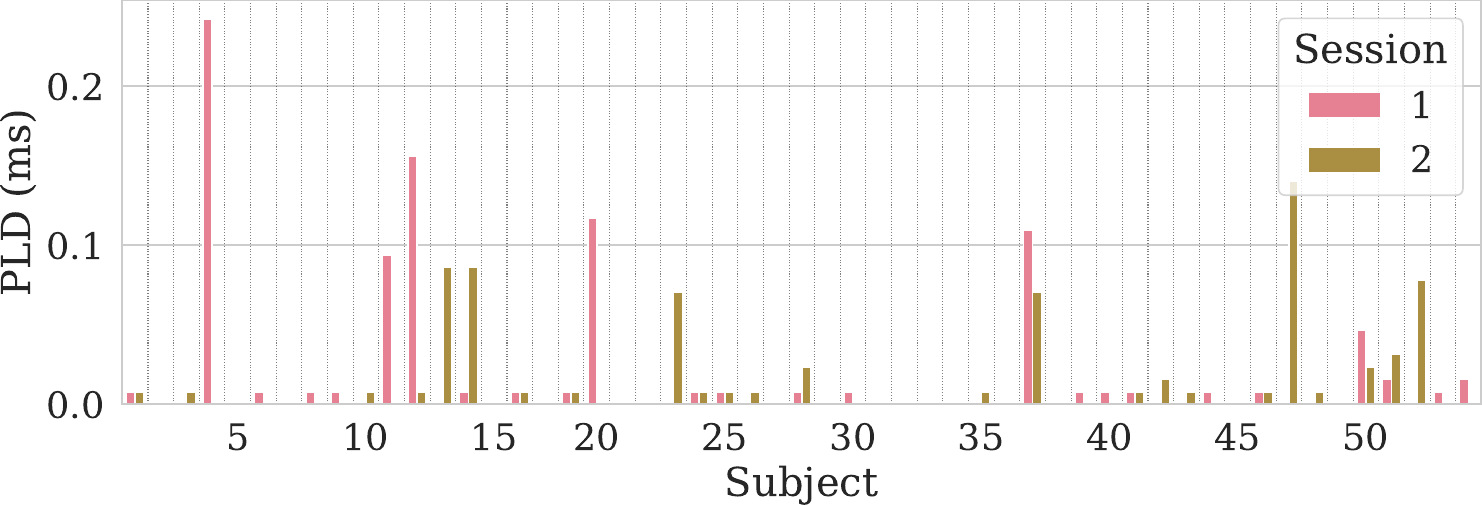}
	\caption{The \textbf{PLD} between the P300 peak latency in channel "O1" in the evoked real and generated data.}
	\label{fig:latency}
\end{figure}

\mysubsection{Peak amplitude and latency:}
The PLD and PAD of the generated data are also much lower compared to the between-session real-data baseline. The average PAD at the best model, according to ABA, is 0.48~\si{\micro\volt}. Using the same model, the maximum PAD is 2.19~\si{\micro\volt}. This can likely be improved upon by optimizing the SNR of the corrector, either using a hyperparameter search or by fine-tuning using a subject- and session-specific approach. Nonetheless, the fact that the between-session PAD is much larger does indicate that the model can generate amplitudes that are at least in line with a particular session and subject. Additionally, the PLD values suggest that the peak position of the generated data is extremely close to that of the real data, with a difference of 0.016~\si{ms} on the same model. This is less than half of the between-session difference. However, there are large inter-subject differences in the PLD of the generated data as can be seen in \autoref{fig:latency}. These can be attributed to the sensitivity of the metric to slight deviations in the generated samples when there are multiple peaks in the selected channel. For example, subject~4 has a high PLD in session~1, however, this does not necessarily mean that the peak of the generated samples is highly dissimilar to the peak in the real data. Instead, the amplitude delta between the highest and the second-highest peak of the real data is quite small, which means that the generated data can have the second-highest peak as the highest peak, thereby disproportionately influencing the PLD. This also influences the PLD of the real data.

\mysubsection{FID performance:}
The FID is mostly useful for future comparisons, as it requires other models that are trained on the same dataset. Our model achieves a FID of $93e^ {-4}$. In the absence of other models, a few baselines are computed to establish a frame of reference. Firstly, the average FID over 20 times computing the FID on two random halves of the real data is $6.90e^{-4}$. Secondly, the FID comparing sessions one and two is $125e^{-4}$. Lastly, the FID comparing the first 26 subjects to the last 25 subjects is $1611e^{-4}$. Thus, it seems that our model can achieve a FID that is slightly better than the FID between sessions one and two but is nowhere close to the FID computed on two random halves of the real data. However, it should be noted that EEGNet is trained on the real data for which subsequently the FID is calculated. Moreover, it has been established that using a conditioned model, instead of the unconditioned model can decrease the FID~\cite{hoClassifierFree}. 


\mysection{Discussion}
\\
In this work, we aimed to create a conditional diffusion model using classifier-free guidance, that does not lose specificity during sampling. The results indicate that the model can indeed create subject-, session-, and class-specific ERP data that is quite similar to the real data of the Lee 2019 ERP dataset. 

\mysubsection{Amplitude, latency, and diversity are well-modeled:} The classifier performance on the real and generated data is highly similar, even for the worst subject. Given that the features on which the classifier is trained are based on amplitude and latency, it is no surprise that the PAD and PLD of the generated data are better than the between-session variability baseline. Furthermore, the diversity, as measured by the SD-MD, also seems to be modeled reasonably well, as it can outperform the between-session variability baseline. 

\mysubsection{Limitation of the PLD:} 
The PLD should be interpreted with caution because it is unreliable when there are multiple peaks of similar height in the data. This can be addressed by either 1) only computing the PLD when there is only one prominent peak or 2) by computing the peaks of multiple subsets (i.e. 80\,\%) of the real data and only using the lowest PLD.

\mysubsection{Potential applications:} 
There are a variety of potential applications for the model presented here. For example, it can be used to alleviate the class imbalance by sampling from the target class. In addition, it can generate additional data to train a classifier, potentially slightly improving the robustness and accuracy of the model. Furthermore, the trained weights of the model can be used for transfer learning, which would allow fine-tuning on a different ERP dataset. Lastly, enlarging datasets with the proposed model may be specifically valuable for the benchmarking of novel algorithms. The diffusion model, however, can not be expected to deliver samples from outside the distribution of the training data.

\mysubsection{Dataset limitations and solutions:}
All in all, the ability of the generated data to achieve comparable performance on most metrics to real data, and the high visual similarity of the covariance matrices and ERPs, are promising results for conditional diffusion models that generate EEG data directly. Nonetheless, it should be noted that the model was trained on a rather large dataset, so it remains to be seen how well the results translate to training on smaller datasets. However, the ability to generate data from specific conditions during sampling, while being trained on a full dataset, might make conditioned models more data-efficient compared to unconditioned models.


\mysubsection{Conclusion:}
In this work, we introduce the first diffusion model that is conditioned on subject, session, and classes using classifier-free guidance, that can generate high-quality EEG data for each condition. This enables training on complete datasets, without losing specificity during sampling. Additionally, we introduce multiple domain-specific metrics that can assist in model evaluation and fine-tuning. This conditional diffusion model can now be used to generate high-quality data for all subjects, sessions, and classes present in the Lee ERP dataset. 

\mysection{Acknowledgements}

We thank Dr. Luca Ambrogioni for his insights into diffusion models.
This work in part is supported by the Donders Center
for Cognition (DCC) and by the project Dutch Brain Interface Initiative
(DBI2), project number 024.005.022 of the research programme Gravitation
which is (partly) financed by the Dutch Research Council (NWO). OnePlanet Research Center acknowledges the support of the Province of Gelderland.




\mysection{References}
\printbibliography[heading=none]

\end{document}